\documentclass{article} % For LaTeX2e
\usepackage{tccml_iclr2025_conference,times}

\usepackage{amsmath,amsfonts,bm}

\def\eqref#1{equation~\ref{#1}}
\def\1{\bm{1}}

\DeclareMathAlphabet{\mathsfit}{\encodingdefault}{\sfdefault}{m}{sl}
\SetMathAlphabet{\mathsfit}{bold}{\encodingdefault}{\sfdefault}{bx}{n}

\usepackage{hyperref}
\usepackage{graphicx}
\usepackage{amsmath,amsfonts,amssymb,amsthm}
\usepackage{enumite m}
\usepackage{tabularx}
\usepackage{multirow}
\usepackage{booktabs}

\title{ClimateChat: Designing Data and Methods for Instruction Tuning LLMs to Answer Climate Change Queries}

\author{
  Zhou Chen \quad
  Xiao Wang \quad
  Yuanhong Liao \quad
  Ming Lin \quad
  Yuqi Bai \\
  Tsinghua University, China \\
  \texttt{chenz22@mails.tsinghua.edu.cn, yuqibai@mail.tsinghua.edu.cn}
}

\iclrfinalcopy % Uncomment for camera-ready version, but NOT for submission.
\usepackage{float}
\begin{document}

\maketitle

\begin{abstract}
As the issue of global climate change becomes increasingly severe, the demand for research in climate science continues to grow. Natural language processing technologies, represented by Large Language Models (LLMs), have been widely applied to climate change-specific research, providing essential information support for decision-makers and the public. Some studies have improved model performance on relevant tasks by constructing climate change-related instruction data and instruction-tuning LLMs. However, current research remains inadequate in efficiently producing large volumes of high-precision instruction data for climate change, which limits further development of climate change LLMs. This study introduces an automated method for constructing instruction data. The method generates instructions using facts and background knowledge from documents and enhances the diversity of the instruction data through web scraping and the collection of seed instructions. Using this method, we constructed a climate change instruction dataset, named ClimateChat-Corpus, which was used to fine-tune open-source LLMs, resulting in an LLM named ClimateChat. Evaluation results show that ClimateChat significantly improves performance on climate change question-and-answer tasks. Additionally, we evaluated the impact of different base models and instruction data on LLM performance and demonstrated its capability to adapt to a wide range of climate change scientific discovery tasks, emphasizing the importance of selecting an appropriate base model for instruction tuning. This research provides valuable references and empirical support for constructing climate change instruction data and training climate change-specific LLMs.
\end{abstract}

\section{Introduction}

As global climate change issues intensify, climate change research has become a focal point worldwide \citep{seddon2022harnessing,abrahms2023climate}. Scientists are committed to exploring new methods and tools to deeply understand and accurately predict the various impacts of climate change and address the challenges it poses \citep{brzic2023detecting,shiwakoti2024analyzing}. Natural Language Processing (NLP) technology has shown great potential in handling climate-related tasks \citep{chen2024preparedllm,callaghan2021machine,ChatClimate,vaghefi2022deep,mallick2024understanding,mallick2024analyzing}.  \citet{Cody2015ClimateCS} utilizes happiness scores from texts to assess the impact of climate change news on public sentiment. However, such methods often neglect the semantic differences of texts in various contexts. The transformer-based BERT language model effectively captures contextual information \citep{BERT}.  \citet{callaghan2021machine} use the BERT language model to identify and categorize observed climate impact studies. ClimateBert improves accuracy in various climate-related tasks through domain-adaptive pre-training on climate-related texts \citep{Climatebert}. \citet{bingler2022cheap} facilitates large-scale analysis of climate disclosures by fine-tuning ClimateBert. \citet{stammbach2022environmental} fine-tunes a RoBERTa to reliably detect environmental claims.

As NLP technology evolves, models are now capable of handling not just traditional tasks like classification but also effectively communicating with humans. The Climate Bot can respond to climate change questions based on document-based evidence \cite{rony2022climate}. \citet{ChatClimate} uses GPT-4 to answer climate change questions based on accurate external knowledge, providing reliable information for decision-makers and the public. Furthermore, instruction data are used to fine-tune LLMs to enhance their performance on climate change tasks.  \citet{Climategpt} involved the manual construction of climate change-related instruction data, a process that required substantial human effort and expert knowledge. \citet{arabic} enables ChatGPT to act as a climate change expert, automating the generation of conversational-style instruction data, thus improving data production efficiency but also presenting the risk of knowledge hallucination.

This study introduces an automated method for constructing large-scale and highly accurate climate change instruction data. This method initially generates instructions and answers using facts and background knowledge from documents, enabling the efficiently creation of a large amount of highly accurate instruction data. It also enriches the diversity of the instruction data and broadens its scale through web scraping and the collection of seed instructions. We use the automated constructed instruction data to fine-tune LLMs and conduct evaluations to demonstrate the effectiveness of this method. We evaluated the impact of different base models and instruction data on LLM performance, and highlighted that using an inappropriate base model for instruction tuning can worsen the occurrence of knowledge hallucination. Additionally, we assessed the effects of instruction tuning and the Retrieval-Augmented Generation (RAG) method on LLM performance, emphasizing the importance of instruction tuning for enhancing LLM.
 
\section{Method}\label{sec:methods}
This section describes the methodology used to construct a climate change-related instruction dataset and to develop an LLM named ClimateChat, which is tailored to climate change question answering.

\subsection{ClimateChat-Corpus}
Referring to the latest advancements in the NLP field, we propose using three strategies to automatically construct climate change-related instruction data: Self-QA, Web Scraping, and Self-Instruct. The constructed dataset is named ClimateChat-Corpus.

\begin{figure*}[h]
\centering
\includegraphics[width=1\textwidth]{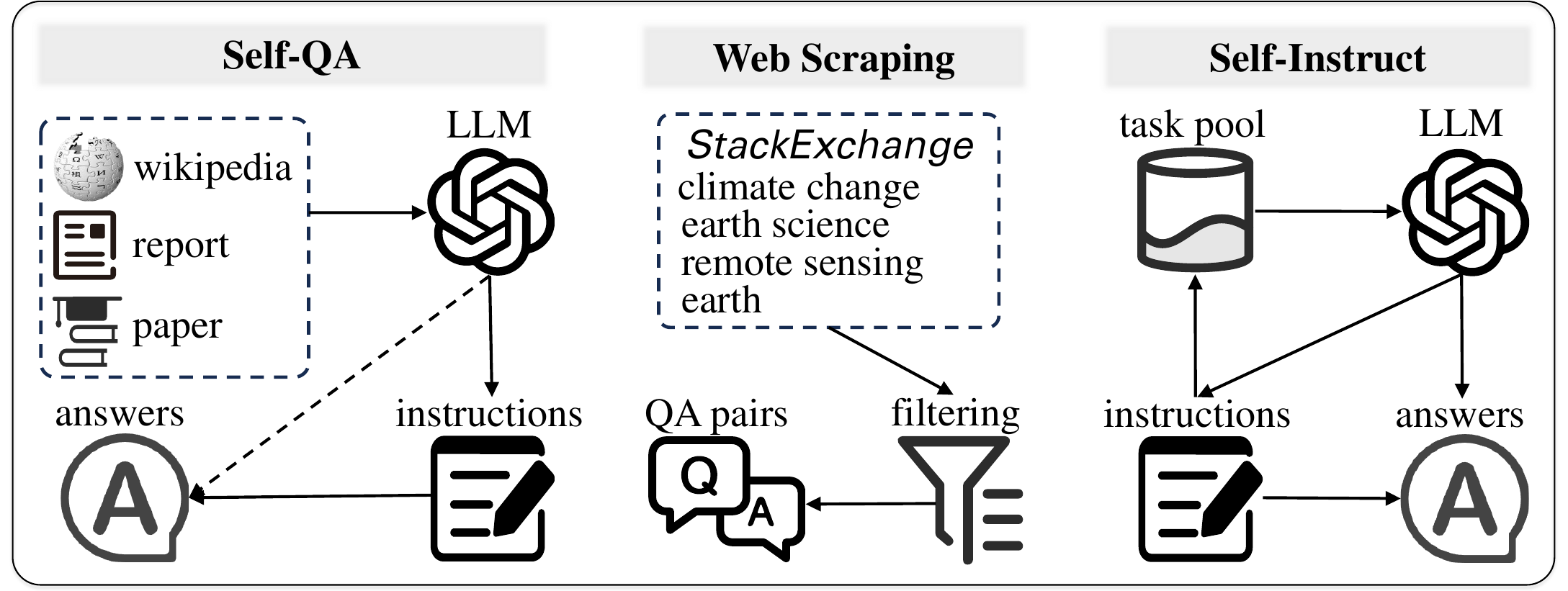}
\caption{Construction of the ClimateChat-Corpus.}
\label{ClimateChat_Corpus}
\end{figure*}

\paragraph{Self-QA}
Initially, extensive materials relating to climate change and earth sciences were gathered, including Wikipedia pages, reports (e.g., The Sixth Assessment Report of the IPCC), and academic papers. We then segmented these texts into smaller paragraphs and used GPT-4 to generate questions based on the facts contained within these paragraphs. Finally, we input both the questions and their corresponding paragraphs into GPT-4 to generate the answers.

\paragraph{Web Scraping}
We collected question-and-answer pairs from StackExchange, focusing on topics such as climate change, earth sciences, remote sensing, and earth. To ensure data balance, only the highly recommended answers for each question were selected. Answers with fewer than three recommendations were excluded to maintain quality.

\paragraph{Self-Instruct}
We amassed a collection of high-quality, climate-related scientific questions from various websites to serve as seed tasks stored in a task pool. Using a few-shot learning approach, tasks were randomly selected as context for GPT-4 to generate instructions and answers. Valid instructions were subsequently added to the task pool to enhance its diversity.

\subsection{Model Training}

\paragraph{Base model selection}
JiuZhou \citep{JiuZhou} is an LLM specifically designed for geoscience. It outperformed GPT-3.5 on the GeoBench \citep{K2} objective tasks by leveraging a substantial corpus of geoscience literature for continued pre-training on Mistral-7B. Given the close relationship between climate change and geoscience, we selected JiuZhou as the base model for ClimateChat. The significance of using JiuZhou as the base model is shown in Section \ref{Result} and further discussed in Appendix \ref{sec:appendixA}.

\paragraph{Training data}
The generalization capability of an LLM is a critical feature. To equip ClimateChat with the ability to handle a variety of climate-related tasks, our training data included both specialized and general datasets. We used the ClimateChat-Corpus as specialized data and selected widely-recognized general datasets such as Alpaca-GPT4 \citep{Alpaca-GPT4}, Dolly \citep{DatabricksBlog2023DollyV2} and BELLE \citep{BELLE}.

\paragraph{Instruction tuning}
LoRA (Low-Rank Adaptation) \citep{lora} is an efficient method for instruction tuning LLMs. By training low-rank matrices and integrating them into the original model, this method enables rapid model adaptation, reducing computational demands.

\section{Result} \label{Result}
ClimateChat was trained using the base model, data, and training methods described in Section \ref{sec:methods}. To further validate the effectiveness of our approach, it was compared against two baseline models developed in this study: (\emph{i}) \textbf{JiuZhouChat}, which was trained using only general instruction data but with the same base model and training methods as ClimateChat; and (\emph{ii}) \textbf{ClimateMistral}, which used the base model of Mistral-7B with identical training data and methods as ClimateChat.

A set of questions and answers was collected from climate change-related disciplines on Wikipedia, and the answer options were expanded using GPT-4 to construct the objective tasks. We used 5-grams to remove data related to the ClimateChat-Corpus. The distribution of questions across the disciplines was as follows: 26 in climate science, 22 in ecology, 18 in environmental science, 20 in health sciences, and 19 in geography. For the subjective tasks, 45 questions related to climate were selected from GeoBench. Responses were evaluated using GPT-4 based on six criteria: helpfulness, relevance, accuracy, depth, creativity, and detail, with each criterion scored on a scale from 1 to 3. Human evaluations are documented in Appendix \ref{sec:appendixA}. Tables \ref{tab:objective} and \ref{tab:subjective} display scores of the models on objective and subjective tasks, respectively. The findings are as follows:

\paragraph{The LLM instruction-tuned with the ClimateChat-Corpus demonstrated significant improvements in performance on climate change question-answering tasks}
ClimateChat consistently outperformed JiuZhouChat, which did not utilize the ClimateChat-Corpus for instruction tuning. This suggests that climate change-related instruction data can enhance the ability of LLMs to leverage internal knowledge to address climate change issues, underscoring the significant value of the ClimateChat-Corpus for training LLMs specialized in climate change question-answering tasks.

\paragraph{Utilizing JiuZhou as the base model significantly boosted performance on climate change question-answering tasks}
The scores of ClimateChat were notably higher than those of ClimateMistral, which did not continue pre-training using geoscience or climate change-related data. This demonstrates that using a domain-specific base model for instruction tuning can substantially enhance the ability of LLMs to solve climate change-related problems. Therefore, selecting JiuZhou as the base model is highly beneficial for training LLMs dedicated to climate change question-answering.

To further underscore the potential of ClimateChat for advancing climate change scientific discovery, its capabilities in scientific literature retrieval and hypothesis formulation are detailed in Appendix \ref{sec:appendixB}. These examples underscore the adaptability of ClimateChat across a broad spectrum of climate change scientific tasks, effectively addressing the diverse needs of researchers.

There is a view that enhancing LLM capabilities using the RAG method can improve its performance in specific domain, even without instruction tuning. In Appendix \ref{sec:appendixC}, we explore the impact of these two methods on LLM performance. The experimental results indicate that, while RAG is effective, combining RAG with an instruction-tuned LLM yields the best performance.

\begin{table}[ht]
\centering
\resizebox{\textwidth}{!}{%
\begin{tabularx}{\linewidth}{l>{\centering\arraybackslash}X>{\centering\arraybackslash}X>{\centering\arraybackslash}X>{\centering\arraybackslash}X>{\centering\arraybackslash}X>{\centering\arraybackslash}X} % X列居中
\hline
\specialrule{0em}{1pt}{1pt}
Model          & Climate & Ecology & Environment & Health & Geography & Average \\ \specialrule{0em}{1pt}{1pt}
\hline
\specialrule{0em}{1pt}{1pt}
ClimateMistral & 76.9\scalebox{0.8}{±3.8}       & 77.3\scalebox{0.8}{±0.0}       & 77.8\scalebox{0.8}{±5.6}           & 70.0\scalebox{0.8}{±0.0}      & 75.3\scalebox{0.8}{±3.2}         & 75.5\scalebox{0.8}{±2.4}       \\
JiuZhouChat    & 81.9\scalebox{0.8}{±2.3}       & 89.5\scalebox{0.8}{±2.7}       & 90.6\scalebox{0.8}{±6.7}           & 81.5\scalebox{0.8}{±6.0}      & 87.9\scalebox{0.8}{±3.2}         & 86.0\scalebox{0.8}{±2.0}       \\ 
ClimateChat    & 93.5\scalebox{0.8}{±2.3}       & 96.8\scalebox{0.8}{±2.7}       & 90.6\scalebox{0.8}{±3.3}           & 90.0\scalebox{0.8}{±8.5}      & 93.2\scalebox{0.8}{±3.2}         & 93.0\scalebox{0.8}{±2.9}       \\ \hline
\end{tabularx}
}
\caption{Scores of baselines and ClimateChat on objective tasks. The evaluation method assesses the accuracy (\%) with which the models answer multiple-choice questions.}
\label{tab:objective}
\end{table}

% \vspace{-10pt}

\begin{table}[ht]
\centering
\resizebox{\textwidth}{!}{%
\begin{tabularx}{\linewidth}{l>{\centering\arraybackslash}X>{\centering\arraybackslash}X>{\centering\arraybackslash}X>{\centering\arraybackslash}X>{\centering\arraybackslash}X>{\centering\arraybackslash}X>{\centering\arraybackslash}X} % X列居中
\hline
\specialrule{0em}{1pt}{1pt}
Model          & Helpfulness & Depth & Accuracy & Relevance & Creativity & Detail & Average \\ 
\specialrule{0em}{1pt}{1pt}
\hline
\specialrule{0em}{1pt}{1pt}
ClimateMistral & 124.5                & 111.0                & 134.0               & 134.0            & 76.0                  & 119.0             & 116.4         \\
JiuZhouChat    & 123.0                  & 102.0                & 130.0               & 133.0            & 75.0                  & 117.0             & 113.5            \\
ClimateChat    & 131.0                  & 129.0                & 134.0               & 134.0            & 97.0                  & 132.0             & 126.2         \\ \hline
\end{tabularx}
}
\caption{Scores of baselines and ClimateChat on subjective tasks. The evaluation method entails scoring the responses of LLMs via GPT-4. Results of human evaluations are provided in Appendix \ref{sec:appendixA}.}
\label{tab:subjective}
\end{table}

% \vspace{-14pt}

\section{Conclusion}

This study proposes a method for automatically constructing climate change-related instruction data. This method can quickly generate a large amount of accurate, specialized instruction data, reducing the burden on climate science experts. We fine-tuned an LLM using the automatically generated instruction data, enhancing its performance in climate change scientific tasks. To further investigate the impact of the base model and instruction data on the climate change capabilities of LLMs, we conducted a series of ablation experiments. The results show that using an LLM with more climate change knowledge as the base model leads to better performance during the instruction tuning phase. Moreover, specialized climate-related instructions augment the ability of LLMs to respond to pertinent questions and assist in retrieving scientific literature and formulating scientific hypotheses. This underscores the adaptability of ClimateChat to various climate-related scientific discovery tasks.

\section{Disccusion}

The automated instruction data generation method outlined in this study includes three strategies: Self-QA, Web Scraping, and Self-Instruct. Self-QA produces instruction data based on highly reliable climate and earth science literature, forming a crucial part of the ClimateChat-Corpus. While this method alone may result in lower diversity, the diversity of instruction data is essential \citep{instag}. Therefore, we employ the Web Scraping and Self-Instruct strategies to expand the instruction data and enhance the diversity of dataset. Notably, although instructions generated by LLMs may include hallucinations, LLMs possess strong judgment capabilities. We utilize LLMs to filter out newly generated instructions, ensuring a robust level of trustworthiness in the instruction data. To the best of our knowledge, accessing open climate change instruction data is challenging, which is why we have not compared ClimateChat-Corpus with human-constructed instruction data. The outstanding performance on objective and subjective demonstrates the effectiveness of our approach. Additionally, we found that while climate change instruction data can enhance LLM performance on related tasks, using an LLM without sufficient knowledge may lead to increased occurrences of hallucinations. Previous studies have also demonstrated this \citep{BERT,gekhman2024}, and we will detail our findings in Appendix \ref{sec:appendixA}.

\section*{Ethics Statement}
We acknowledge that the automated generation of instruction data may inadvertently introduce biases. To address this, we have implemented robust measures during the evaluation and deployment processes to identify and mitigate potential issues. ClimateChat is designed as a supportive tool for researchers, policymakers, and the public, with a steadfast commitment to promoting equitable access to its capabilities, especially for underrepresented communities in climate research. By fostering collaboration and deepening the understanding of climate science, ClimateChat strives to make a meaningful contribution to tackling the challenges of climate change.

\section*{Data availability}

The ClimateChat model checkpoint can be accessed at \url{https://huggingface.co/itpossible/ClimateChat}. We used the training code available at \url{https://github.com/THU-ESIS/JiuZhou}.

\section*{Acknowledgements}

This study was supported by the Major Program of the National Natural Science Foundation of China (Grant Number: 42090015).


\bibliography{iclr2025_conference}

\begin{thebibliography}{26}
\providecommand{\natexlab}[1]{#1}
\providecommand{\url}[1]{\texttt{#1}}
\expandafter\ifx\csname urlstyle\endcsname\relax
  \providecommand{\doi}[1]{doi: #1}\else
  \providecommand{\doi}{doi: \begingroup \urlstyle{rm}\Url}\fi

\bibitem[Abrahms et~al.(2023)Abrahms, Carter, Clark-Wolf, Gaynor, Johansson, McInturff, Nisi, Rafiq, and West]{abrahms2023climate}
Briana Abrahms, Neil~H Carter, TJ~Clark-Wolf, Kaitlyn~M Gaynor, Erik Johansson, Alex McInturff, Anna~C Nisi, Kasim Rafiq, and Leigh West.
\newblock Climate change as a global amplifier of human--wildlife conflict.
\newblock \emph{Nature Climate Change}, 13\penalty0 (3):\penalty0 224--234, 2023.
\newblock \doi{10.1038/s41558-023-01608-5}.

\bibitem[Bingler et~al.(2022)Bingler, Kraus, Leippold, and Webersinke]{bingler2022cheap}
Julia~Anna Bingler, Mathias Kraus, Markus Leippold, and Nicolas Webersinke.
\newblock Cheap talk in corporate climate commitments: The role of active institutional ownership, signaling, materiality, and sentiment.
\newblock Technical report, Swiss Finance Institute, 2022.
\newblock URL \url{https://EconPapers.repec.org/RePEc:chf:rpseri:rp2201}.

\bibitem[Brzic et~al.(2023)Brzic, Boticki, and Bagic~Babac]{brzic2023detecting}
Barbara Brzic, Ivica Boticki, and Marina Bagic~Babac.
\newblock Detecting deception using natural language processing and machine learning in datasets on covid-19 and climate change.
\newblock \emph{Algorithms}, 16\penalty0 (5):\penalty0 221, 2023.
\newblock \doi{10.1016/j.nlp.2024.100057}.

\bibitem[Callaghan et~al.(2021)Callaghan, Schleussner, Nath, Lejeune, Knutson, Reichstein, Hansen, Theokritoff, Andrijevic, Brecha, et~al.]{callaghan2021machine}
Max Callaghan, Carl-Friedrich Schleussner, Shruti Nath, Quentin Lejeune, Thomas~R Knutson, Markus Reichstein, Gerrit Hansen, Emily Theokritoff, Marina Andrijevic, Robert~J Brecha, et~al.
\newblock Machine-learning-based evidence and attribution mapping of 100,000 climate impact studies.
\newblock \emph{Nature climate change}, 11\penalty0 (11):\penalty0 966--972, 2021.
\newblock \doi{10.1038/s41558-021-01168-6}.

\bibitem[Chen et~al.(2024)Chen, Lin, Wang, Zang, and Bai]{chen2024preparedllm}
Zhou Chen, Ming Lin, Zimeng Wang, Mingrun Zang, and Yuqi Bai.
\newblock Preparedllm: Effective pre-pretraining framework for domain-specific large language models.
\newblock \emph{Big Earth Data}, 8\penalty0 (4):\penalty0 649--672, 2024.
\newblock \doi{10.1080/20964471.2024.2396159}.

\bibitem[Chen et~al.(2025)Chen, Lin, Zang, Wang, and Bai]{JiuZhou}
Zhou Chen, Ming Lin, Mingrun Zang, Zimeng Wang, and Yuqi Bai.
\newblock Jiuzhou: Open foundation language models and effective pre-training framework for geoscience.
\newblock \emph{International Journal of Digital Earth}, 18\penalty0 (1):\penalty0 1--30, 2025.
\newblock \doi{10.1080/17538947.2025.2449708}.

\bibitem[Cody et~al.(2015)Cody, Reagan, Mitchell, Dodds, and Danforth]{Cody2015ClimateCS}
Emily~M. Cody, Andrew~J. Reagan, Lewis Mitchell, Peter~Sheridan Dodds, and Christopher~M. Danforth.
\newblock Climate change sentiment on twitter: An unsolicited public opinion poll.
\newblock \emph{PLoS ONE}, 10, 2015.
\newblock URL \url{https://api.semanticscholar.org/CorpusID:872939}.

\bibitem[Conover et~al.(2023)Conover, Hayes, Mathur, Xie, Wan, Shah, Ghodsi, Wendell, Zaharia, and Xin]{DatabricksBlog2023DollyV2}
Mike Conover, Matt Hayes, Ankit Mathur, Jianwei Xie, Jun Wan, Sam Shah, Ali Ghodsi, Patrick Wendell, Matei Zaharia, and Reynold Xin.
\newblock Free dolly: Introducing the world's first truly open instruction-tuned llm, 2023.
\newblock URL \url{https://www.databricks.com/blog/2023/04/12/dolly-first-open-commercially-viable-instruction-tuned-llm}.

\bibitem[Deng et~al.(2024)Deng, Zhang, He, Chen, Shi, Xu, Fu, Zhang, Wang, Zhou, Lin, and He]{K2}
Cheng Deng, Tianhang Zhang, Zhongmou He, Qiyuan Chen, Yuanyuan Shi, Yi~Xu, Luoyi Fu, Weinan Zhang, Xinbing Wang, Chenghu Zhou, Zhouhan Lin, and Junxian He.
\newblock K2: A foundation language model for geoscience knowledge understanding and utilization.
\newblock WSDM '24, pp.\  161–170, New York, NY, USA, 2024. Association for Computing Machinery.
\newblock ISBN 9798400703713.
\newblock \doi{10.1145/3616855.3635772}.
\newblock URL \url{https://doi.org/10.1145/3616855.3635772}.

\bibitem[Devlin et~al.(2019)Devlin, Chang, Lee, and Toutanova]{BERT}
Jacob Devlin, Ming-Wei Chang, Kenton Lee, and Kristina Toutanova.
\newblock Bert: Pre-training of deep bidirectional transformers for language understanding.
\newblock In \emph{NAACL}, pp.\  1--16, 2019.
\newblock \doi{10.18653/v1/N19-1423}.

\bibitem[Gekhman et~al.(2024)Gekhman, Yona, Aharoni, Eyal, Feder, Reichart, and Herzig]{gekhman2024}
Zorik Gekhman, Gal Yona, Roee Aharoni, Matan Eyal, Amir Feder, Roi Reichart, and Jonathan Herzig.
\newblock Does fine-tuning llms on new knowledge encourage hallucinations?, 2024.
\newblock URL \url{https://arxiv.org/abs/2405.05904}.

\bibitem[Hu et~al.(2022)Hu, Shen, Wallis, Allen-Zhu, Li, Wang, Wang, and Chen]{lora}
Edward~J. Hu, Yelong Shen, Phillip Wallis, Zeyuan Allen-Zhu, Yuanzhi Li, Shean Wang, Lu~Wang, and Weizhu Chen.
\newblock Lora: Low-rank adaptation of large language models.
\newblock In \emph{The Eleventh International Conference on Learning Representations}, pp.\  1--26, 2022.
\newblock \doi{10.48550/arXiv.2106.09685}.

\bibitem[Ji et~al.(2023)Ji, Deng, Gong, Peng, Niu, Zhang, Ma, and Li]{BELLE}
Yunjie Ji, Yong Deng, Yan Gong, Yiping Peng, Qiang Niu, Lei Zhang, Baochang Ma, and Xiangang Li.
\newblock Exploring the impact of instruction data scaling on large language models: An empirical study on real-world use cases, 2023.
\newblock URL \url{https://arxiv.org/abs/2303.14742}.

\bibitem[Lu et~al.(2023)Lu, Yuan, Yuan, Lin, Lin, Tan, Zhou, and Zhou]{instag}
Keming Lu, Hongyi Yuan, Zheng Yuan, Runji Lin, Junyang Lin, Chuanqi Tan, Chang Zhou, and Jingren Zhou.
\newblock \# instag: Instruction tagging for analyzing supervised fine-tuning of large language models.
\newblock In \emph{The Twelfth International Conference on Learning Representations}, 2023.
\newblock \doi{10.48550/arXiv.2308.07074}.

\bibitem[Mallick et~al.(2024{\natexlab{a}})Mallick, Bergerson, Verner, Hutchison, Levy, and Balaprakash]{mallick2024understanding}
Tanwi Mallick, Joshua~David Bergerson, Duane~R Verner, John~K Hutchison, Leslie-Anne Levy, and Prasanna Balaprakash.
\newblock Understanding the impact of climate change on critical infrastructure through nlp analysis of scientific literature.
\newblock \emph{Sustainable and Resilient Infrastructure}, pp.\  1--18, 2024{\natexlab{a}}.
\newblock \doi{10.1080/23789689.2024.2355772}.

\bibitem[Mallick et~al.(2024{\natexlab{b}})Mallick, Murphy, Bergerson, Verner, Hutchison, and Levy]{mallick2024analyzing}
Tanwi Mallick, John Murphy, Joshua~David Bergerson, Duane~R Verner, John~K Hutchison, and Leslie-Anne Levy.
\newblock Analyzing regional impacts of climate change using natural language processing techniques.
\newblock \emph{arXiv preprint arXiv:2401.06817}, 2024{\natexlab{b}}.
\newblock \doi{10.48550/arXiv.2401.06817}.

\bibitem[Mullappilly et~al.(2023)Mullappilly, Shaker, Thawakar, Cholakkal, Anwer, Khan, and Khan]{arabic}
Sahal Mullappilly, Abdelrahman Shaker, Omkar Thawakar, Hisham Cholakkal, Rao Anwer, Salman Khan, and Fahad Khan.
\newblock {A}rabic mini-{C}limate{GPT} : A climate change and sustainability tailored {A}rabic {LLM}.
\newblock In Houda Bouamor, Juan Pino, and Kalika Bali (eds.), \emph{Findings of the Association for Computational Linguistics: EMNLP 2023}, pp.\  14126--14136, Singapore, December 2023. Association for Computational Linguistics.
\newblock \doi{10.18653/v1/2023.findings-emnlp.941}.
\newblock URL \url{https://aclanthology.org/2023.findings-emnlp.941}.

\bibitem[Peng et~al.(2023)Peng, Li, He, Galley, and Gao]{Alpaca-GPT4}
Baolin Peng, Chunyuan Li, Pengcheng He, Michel Galley, and Jianfeng Gao.
\newblock Instruction tuning with gpt-4, 2023.
\newblock URL \url{https://arxiv.org/abs/2304.03277}.

\bibitem[Rony et~al.(2022)Rony, Zuo, Kovriguina, Teucher, and Lehmann]{rony2022climate}
Md~Rashad Al~Hasan Rony, Ying Zuo, Liubov Kovriguina, Roman Teucher, and Jens Lehmann.
\newblock Climate bot: A machine reading comprehension system for climate change question answering.
\newblock In \emph{IJCAI}, pp.\  5249--5252, 2022.
\newblock \doi{10.24963/ijcai.2022/729}.

\bibitem[Seddon(2022)]{seddon2022harnessing}
Nathalie Seddon.
\newblock Harnessing the potential of nature-based solutions for mitigating and adapting to climate change.
\newblock \emph{Science}, 376\penalty0 (6600):\penalty0 1410--1416, 2022.
\newblock \doi{10.1126/science.abn9668}.

\bibitem[Shiwakoti et~al.(2024)Shiwakoti, Thapa, Rauniyar, Shah, Bhandari, and Naseem]{shiwakoti2024analyzing}
Shuvam Shiwakoti, Surendrabikram Thapa, Kritesh Rauniyar, Akshyat Shah, Aashish Bhandari, and Usman Naseem.
\newblock Analyzing the dynamics of climate change discourse on twitter: A new annotated corpus and multi-aspect classification.
\newblock In \emph{Proceedings of the 2024 Joint International Conference on Computational Linguistics, Language Resources and Evaluation (LREC-COLING 2024)}, pp.\  984--994, 2024.
\newblock URL \url{https://aclanthology.org/2024.lrec-main.88}.

\bibitem[Stammbach et~al.(2022)Stammbach, Webersinke, Bingler, Kraus, and Leippold]{stammbach2022environmental}
Dominik Stammbach, Nicolas Webersinke, Julia Bingler, Mathias Kraus, and Markus Leippold.
\newblock Environmental claim detection.
\newblock \emph{Available at SSRN 4207369}, 2022.
\newblock \doi{10.18653/v1/2023.acl-short.91}.

\bibitem[Thulke et~al.(2024)Thulke, Gao, Pelser, Brune, Jalota, Fok, Ramos, van Wyk, Nasir, Goldstein, et~al.]{Climategpt}
David Thulke, Yingbo Gao, Petrus Pelser, Rein Brune, Rricha Jalota, Floris Fok, Michael Ramos, Ian van Wyk, Abdallah Nasir, Hayden Goldstein, et~al.
\newblock Climategpt: Towards ai synthesizing interdisciplinary research on climate change.
\newblock \emph{arXiv preprint arXiv:2401.09646}, 2024.
\newblock \doi{10.48550/arXiv.2401.09646}.

\bibitem[Vaghefi et~al.(2022)Vaghefi, Muccione, Huggel, Khashehchi, and Leippold]{vaghefi2022deep}
Saeid Vaghefi, Veruska Muccione, Christian Huggel, Hamed Khashehchi, and Markus Leippold.
\newblock Deep climate change: A dataset and adaptive domain pre-trained language models for climate change related tasks.
\newblock In \emph{NeurIPS 2022 Workshop on Tackling Climate Change with Machine Learning}, 2022.
\newblock URL \url{https://www.climatechange.ai/papers/neurips2022/27}.

\bibitem[Vaghefi et~al.(2023)Vaghefi, Stammbach, Muccione, Bingler, Ni, Kraus, Allen, Colesanti-Senni, Wekhof, Schimanski, et~al.]{ChatClimate}
Saeid~Ashraf Vaghefi, Dominik Stammbach, Veruska Muccione, Julia Bingler, Jingwei Ni, Mathias Kraus, Simon Allen, Chiara Colesanti-Senni, Tobias Wekhof, Tobias Schimanski, et~al.
\newblock Chatclimate: Grounding conversational ai in climate science.
\newblock \emph{Communications Earth \& Environment}, 4\penalty0 (1):\penalty0 480, 2023.
\newblock \doi{10.1038/s43247-023-01084-x}.

\bibitem[Webersinke et~al.(2021)Webersinke, Kraus, Bingler, and Leippold]{Climatebert}
Nicolas Webersinke, Mathias Kraus, Julia~Anna Bingler, and Markus Leippold.
\newblock Climatebert: A pretrained language model for climate-related text.
\newblock \emph{arXiv preprint arXiv:2110.12010}, 2021.
\newblock \doi{10.48550/arXiv.2110.12010}.

\end{thebibliography}
\bibliographystyle{iclr2025_conference}

\clearpage  

\appendix

\section{Instruction tuning requires a suitable foundational LLM}\label{sec:appendixA}

\subsection{Experiment and result}

A manual evaluation was conducted on the responses generated by ClimateMistral, JiuZhouChat, and ClimateChat to 45 climate change-related questions from GeoBench to assess response quality. Response quality was categorized into four levels: Significant Inaccuracy, Moderate Inaccuracy, Slight Inaccuracy, and Accurate.

\paragraph{Significant Inaccuracy}
The response contains severe factual errors or fabricated information that fundamentally contradicts known facts or logic, potentially leading to misunderstandings or incorrect decisions. For example, the model misrepresents the timing of events, key figures, or provides entirely incorrect scientific explanations.

\paragraph{Moderate Inaccuracy} The response includes noticeable errors or partially inaccurate information. While these errors may not be completely misleading, they could cause some degree of misunderstanding. For instance, the model provides a partially correct explanation but omits important details or includes some inaccurate statements.

\paragraph{Slight Inaccuracy} The response is generally correct but may contain minor inaccuracies or slightly awkward logic. These flaws are unlikely to significantly affect overall understanding. For example, the response is mostly accurate but may suffer from slightly complex sentence structures or imprecise word choices, affecting clarity.

\paragraph{Accuracy} The response is entirely correct, with accurate information, clear logic, and fluent expression, fully aligned with facts or established scientific principles. For example, the model provides an accurate and detailed answer, including relevant background information and precise references, with appropriate language use.

Table \ref{tab:hull} presents the manual evaluation results for ClimateMistral, JiuZhouChat, and ClimateChat. As shown in the table, ClimateMistral exhibits the highest level of hallucinations, followed by JiuZhouChat, with ClimateChat demonstrating the lowest level of inaccuracies.

\begin{table}[ht]
\centering
\resizebox{\textwidth}{!}{%
\begin{tabular}{lcccc}
\hline
\specialrule{0em}{1pt}{1pt}
Model & Significant Inaccuracy & Moderate Inaccuracy & Slight Inaccuracy & Accuracy \\
\specialrule{0em}{1pt}{1pt}
\hline
\specialrule{0em}{1pt}{1pt}
ClimateMistral & 11 & 18 & 10 & 6 \\
JiuZhouChat    & 3  & 3  & 10 & 29 \\
ClimateChat    & 1  & 4  & 6  & 34 \\
\hline
\end{tabular}
}
\caption{Evaluation results for subjective tasks of baselines and ClimateChat.}
\label{tab:hull}
\end{table}

\textbf{ClimateMistral} This response attempts to quantify sea level rise by listing several factors, including ice melt, thermal expansion, changes in precipitation, and shifts in ocean circulation patterns. However, the specific figures provided (e.g. 35\%, 25\%, 16\%) are not universally recognized by the scientific community, suggesting a knowledge hallucination issue. Such figures could potentially mislead readers.

\textbf{JiuZhouChat} This response provides a broad overview of sea level changes, covering factors such as changes in ocean volume, gravitational shifts, precipitation, and evaporation. However, the content is overly broad and lacks focus on the primary drivers recognized by the scientific community, such as thermal expansion and ice melt.

\textbf{ClimateChat} This response focuses on two main quantitative factors: the thermal expansion of seawater and glacier melting. It clearly states that global temperature increases lead to seawater expansion, and that ice sheet melting (particularly in Antarctica and Greenland) is a major factor in sea level rise. This response directly addresses the question and aligns with the current scientific consensus.

\subsection{Case Analysis}

\begin{figure*}[h]
\centering
\includegraphics[width=1\textwidth]{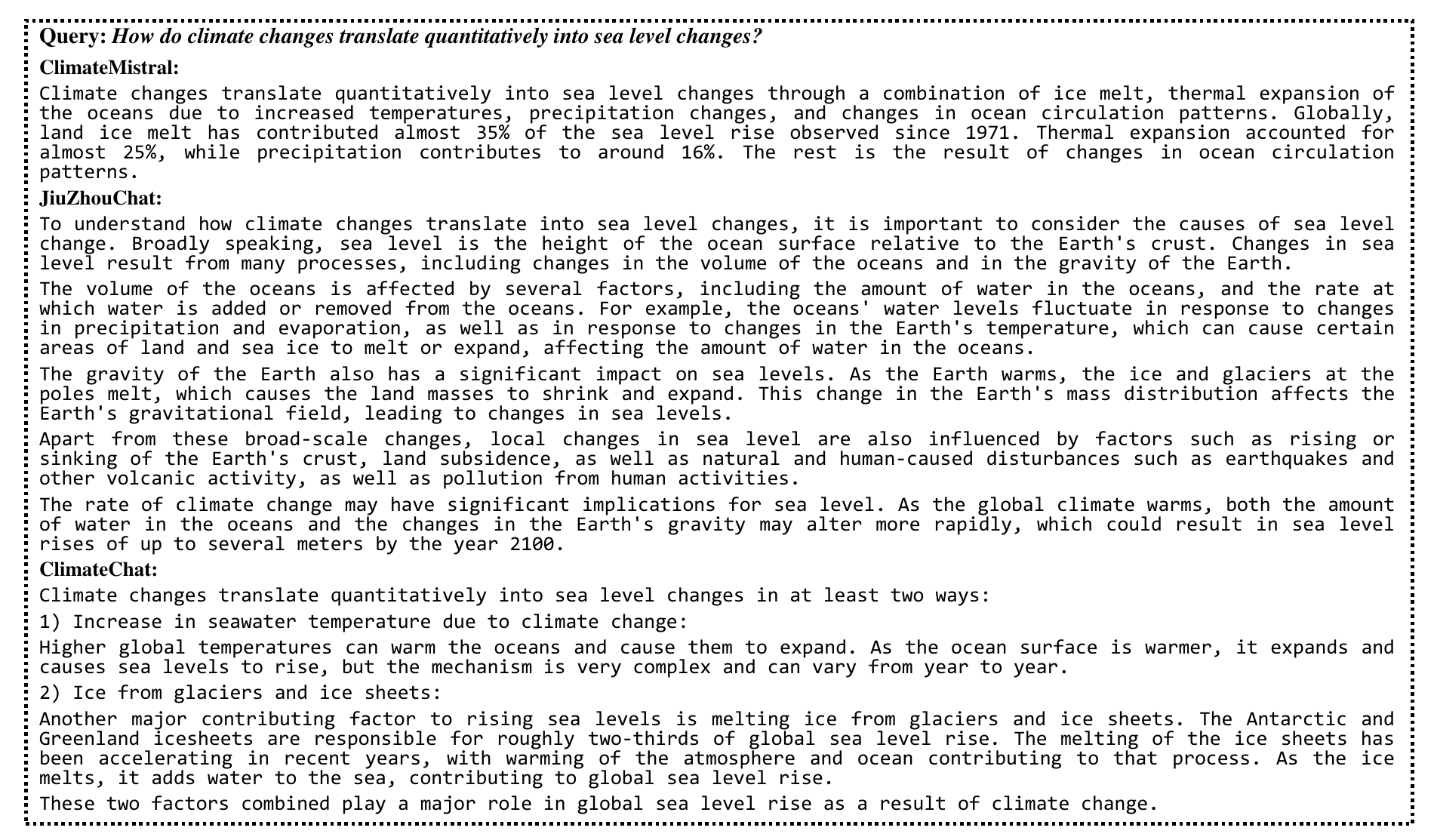}
\caption{Responses of ClimateMistral, JiuZhouChat, and ClimateChat to the climate change question.}
\label{ClimateChat_Case}
\end{figure*}

A climate change-related query was randomly selected from GeoBench: "How do climate changes translate quantitatively into sea level changes?" Subsequently, three models (ClimateMistral, JiuZhouChat, and ClimateChat) were tasked with generating responses to this query. Figure \ref{ClimateChat_Case} depicts the specific responses of each model.

Performance of the three models across different criteria are as follows:

\textbf{Accuracy and Scientific Rigor} The response from ClimateMistral lacks clear scientific substantiation, potentially misleading readers with its inaccurate figures. The response of JiuZhou overlooks the primary drivers of global sea level change. In contrast, the response of ClimateChat aligns most closely with existing scientific research, accurately focusing on the key drivers of sea level rise: thermal expansion and ice melt.

\textbf{Conciseness and Clarity} Although ClimateMistral provides specific data, the absence of scientific substantiation reduces the clarity of its response. The response of JiuZhouChat, while detailed, omits crucial information, potentially confusing readers. The response of ClimateChat is highly relevant, succinctly explaining the translation of climate change into sea level rise via thermal expansion and ice melt.

\textbf{Relevance} While response of ClimateMistral is related to the question, it lacks scientific support. The response of JiuZhouChat covers a wide range of topics, but many of the factors mentioned are not directly relevant or are not of significant importance on a global scale. The response of ClimateChat has the highest relevance to the question, clearly listing and explaining how climate change translates into sea level rise through thermal expansion and ice melt.

Overall, ClimateChat provides the highest quality answer. It most clearly and accurately explains how climate changes quantitatively translate into sea level changes, in alignment with the current scientific consensus. Although ClimateMistral offers quantitative information, its reliability is questionable due to the issue of knowledge hallucination. Despite its comprehensiveness, JiuZhouChat does not effectively address the core of the question.

\subsection{Conclusion}
\textbf{LLMs lacking sufficient climate change knowledge are unsuitable for instruction tuning} Although ClimateMistral was instruction-tuned with the ClimateChat-Corpus, it can provide specific data that, however, lacks a scientific basis. This shortfall is attributed to  inadequate acquisition of climate change knowledge of Mistral during its pre-training phase. Utilizing such instruction data for tuning may yield responses misaligned with factual information, thereby exacerbating knowledge hallucination issues.

\textbf{Instruction tuning with climate change-specific data effectively enhances performance in LLMs possessing adequate knowledge of the topic} Although instruction-tuned with general data, JiuZhouChat provides valuable insights but tends to be verbose and lacks focus on the core issue. This deficiency stems from the failure of JiuZhouChat to learn precise response mechanisms during the instruction tuning phase. Given that JiuZhou is the base model of ClimateChat, which not only possesses comprehensive climate change knowledge but also mastered high-quality responses during instruction tuning, it provides the clearest and most accurate answers.

\section{Ability of ChatClimate for Assisting Scientific Discovery}\label{sec:appendixB}

\subsection{Scientific Literature Retrieval}
Traditional methods of scientific literature retrieval necessitate that researchers generate keywords based on their informational needs to search through documents and subsequently filter the literature to identify articles meriting detailed review. This process is both time-consuming and labor-intensive, proving challenging for newcomers in the research field. To address this, several tools (e.g., Semantic Scholar) were integrated within the system prompt of ClimateChat, enabling autonomous scientific literature retrieval. An illustration of how ClimateChat assists scientists in retrieving climate change-related scientific literature is displayed in the left panel of Figure \ref{fig:scientific_cases}. The need of scientist was to obtain literature on the impacts of sea level rise on coastal cities. The model deduced the need to utilize the Semantic Scholar API for document retrieval and summarized key topics such as 'sea level', 'city', and 'impact' for the literature search. Initially, the model saved the top 20 papers returned by the API, subsequently performing a secondary selection by meticulously reviewing the titles and abstracts. Ultimately, the model selected four pertinent papers, presenting these and their main content to the scientist.

\subsection{Scientific Hypothesis Formulation}
\begin{figure*}[h]
\centering
\includegraphics[width=1\textwidth]{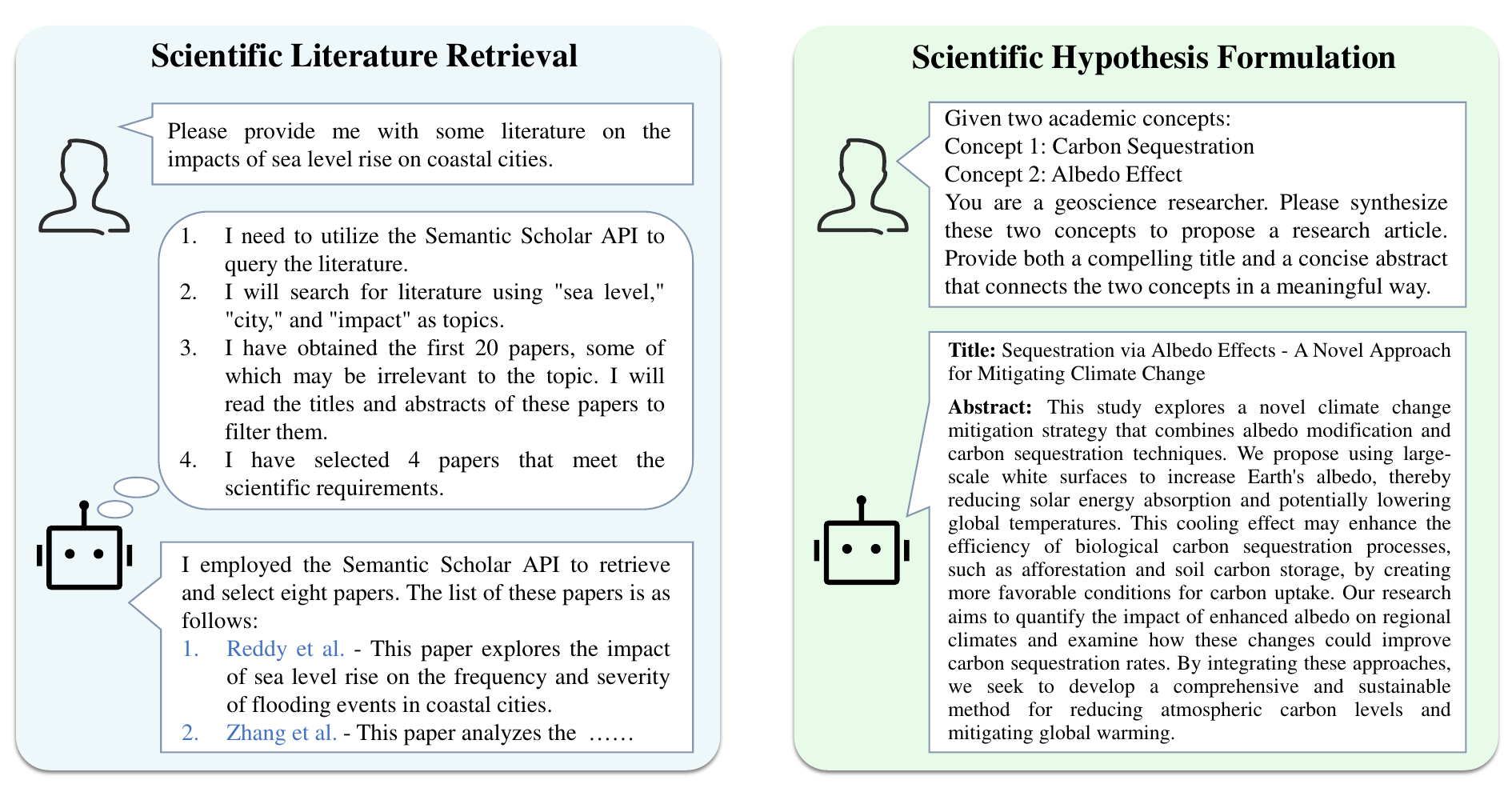}
\caption{Examples of ClimateChat facilitating scientific discovery for scientists. The left panel demonstrates its application in scientific literature retrieval, while the right panel showcases its role in scientific hypothesis formulation.}
\label{fig:scientific_cases}
\end{figure*}

The development of scientific theories typically originates from hypotheses, necessitating that researchers think beyond conventional frameworks and make bold conjectures. During its pre-training and instruction tuning, ClimateChat has assimilated knowledge from millions of scientific papers related to climate change, enabling rapid generation of numerous scientific hypotheses and thereby accelerating discovery. The right panel of Figure \ref{fig:scientific_cases} illustrates how ClimateChat assists scientists in generating scientific hypotheses related to climate change. The scientist required integration of the concepts of Carbon Sequestration and the Albedo Effect. ClimateChat generated article titles and abstracts to systematically deduce how these concepts could inform new research directions. It is important to recognize that hypotheses generated by ClimateChat may be falsified during experimental validation, a process of equal value. Furthermore, incorrect hypotheses reveal the limitations of existing knowledge, thereby guiding future research towards new directions. Scientific hypotheses propel innovation and progress, and the distinctive advantage of ClimateChat resides in its capacity to rapidly generate numerous hypotheses, providing a variety of references for scientific exploration.

\section{Exploring Whether RAG Can Replace Instruction Tuning}\label{sec:appendixC}

\subsection{Background}
The RAG method incorporates an external retrieval mechanism to extract information from knowledge bases, thereby supplementing the background knowledge used for LLM text generation. Through this mechanism, LLMs can generate responses not only based on the knowledge acquired during training, but also by retrieving up-to-date, domain-specific knowledge from external resources, which enhances the accuracy and relevance of their outputs.

In recent years, RAG has emerged as an effective technique for enhancing LLM performance. However, whether RAG can replace instruction tuning, especially for domain-specific tasks such as climate change, remains a question worth exploring. This section explores whether RAG can function as a replacement for instruction tuning, or whether the two methods can operate independently, or perhaps be combined for optimal performance.

\subsection{Experiment and Results}
For the RAG method, we utilized the IPCC (Intergovernmental Panel on Climate Change) reports and Wikipedia as external knowledge bases, which were converted into vector space representations using an embedding model. Climate scientist queries were similarly transformed into vectors, and their similarity to the external knowledge base was calculated. The top five chunks with the highest similarity were concatenated with the query and provided to the LLM to assist in generating responses.
ClimateChat refers to the model trained through instruction tuning with climate-specific data, while JiuZhouChat is the model that was not subjected to climate-specific instruction tuning. By comparing the performance of these two models with and without the RAG method on climate change tasks, we aim to assess whether RAG can replace instruction tuning. The experimental results are shown in Table \ref{tab:RAG}.

\begin{table}[ht]
\centering
\begin{tabular}{c|ccc|ccc}
\hline
\specialrule{0em}{1pt}{1pt}
\multirow{2}{*}{Model} & \multicolumn{3}{c}{w/o RAG}               & \multicolumn{3}{|c}{w/ RAG}                \\ \specialrule{0em}{1pt}{1pt} \cline{2-7} \specialrule{0em}{1pt}{1pt}
                               & Obj. Task & Subj. Task & Average    & Obj. Task & Subj. Task & Average    \\ \specialrule{0em}{1pt}{1pt} \hline \specialrule{0em}{1pt}{1pt}
JiuZhouChat                    & 86.0\scalebox{0.8}{±2.0}       & 113.5           & 99.8\scalebox{0.8}{±1.0}   & 93.2\scalebox{0.8}{±1.9}     & 120.3           & 106.8\scalebox{0.8}{±1.0}  \\
ClimateChat                    & 93.0\scalebox{0.8}{±2.9}       & 126.2           & 109.6\scalebox{0.8}{±1.4}  & 93.8\scalebox{0.8}{±1.7}     & 129.6           & 111.7\scalebox{0.8}{±0.9}  \\ \specialrule{0em}{1pt}{1pt} \hline
\end{tabular}
\caption{Performance comparison of different models with and without RAG.}
\label{tab:RAG}
\end{table}

Table \ref{tab:RAG} demonstrates that the RAG method enables JiuZhouChat to achieve performance levels comparable to ClimateChat on objective tasks. However, a substantial gap persists between the two LLMs on subjective tasks. This indicates that although the RAG method markedly enhances LLM performance on objective tasks, it remains less effective than instruction tuning for subjective tasks. This limitation arises because RAG, by introducing external factual knowledge, enriches the background information accessible to LLMs when addressing queries, but it does not improve the model’s deeper comprehension of the climate change domain. In contrast, instruction tuning effectively activates the climate-specific knowledge acquired by LLMs during pretraining, thereby enhancing the ability of these to understand and address the complex issues within the climate change domain.

Based on the aforementioned experiments and analysis, it is evident that although the RAG method enhances LLM performance on objective tasks, it remains less effective than instruction tuning on subjective tasks. Consequently, RAG cannot substitute for instruction tuning. However, the combination of RAG and instruction tuning significantly augments the capabilities of LLM, outperforming the individual methods.

\subsection{Effectiveness of Combining RAG and Instruction Tuning}

\begin{figure*}[h]
\centering
\includegraphics[width=0.98\textwidth]{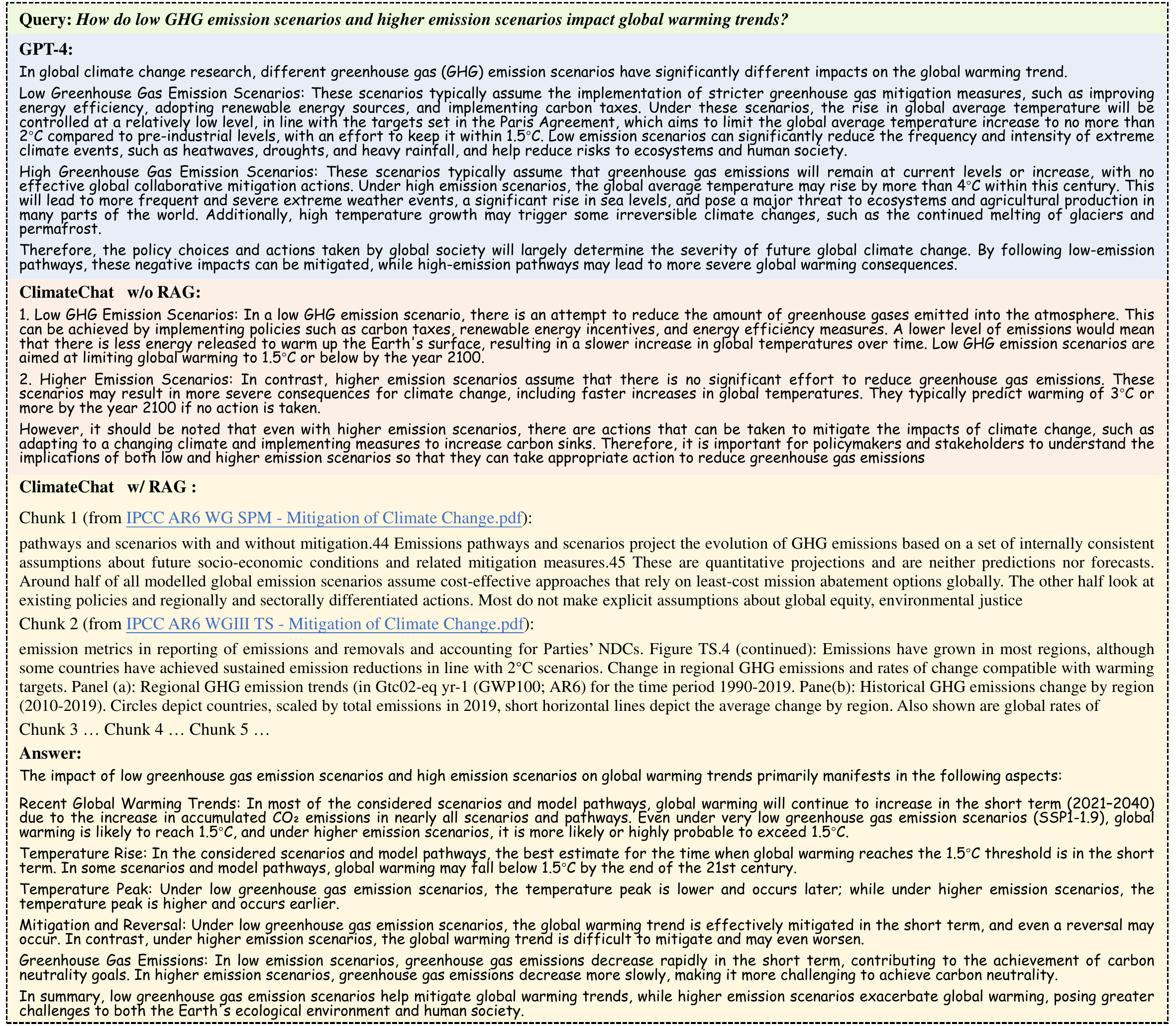}
\caption{Performance of ClimateChat with and without the RAG method versus GPT-4 in answering climate change query.}
\label{fig:RAG}
\end{figure*}

We illustrate the benefits of integrating the RAG and instruction tuning methods in answering climate change-related questions through a case study. Figure \ref{fig:RAG} illustrates the performance differences between ClimateChat with and without the RAG method, in comparison to GPT-4.

In comparison to the response of GPT-4, the RAG-based ClimateChat outperforms in terms of detailed presentation and temporal framework analysis. It not only provides a more comprehensive analysis of short-term (2021–2040) and long-term temperature trends under various emission scenarios, but also specifically highlights that under low-emission scenarios, the temperature peak is lower and occurs later, whereas under high-emission scenarios, the temperature peak is higher and occurs earlier. Furthermore, the RAG-based ClimateChat emphasizes that low-emission scenarios help mitigate global warming and may even lead to a temperature decline, whereas high-emission scenarios pose significant challenges to global warming mitigation efforts. It also underscores that achieving carbon neutrality is more challenging, particularly under high-emission scenarios, due to the slower pace of emission reductions, thereby making the carbon neutrality target more elusive. Overall, LLMs that integrate instruction tuning with RAG methods provide more detailed, complex scenario analyses and policy recommendations, rendering them more practical and forward-looking in addressing climate change queries.

\end{document}